\documentclass[10pt]{report}
\usepackage[authoryear,sort&compress,round]{natbib}
\usepackage{nicefrac}
\usepackage{multirow}
\usepackage{tabularx}
\usepackage{placeins}
\usepackage[nameinlink]{cleveref}

\crefname{equation}{Eq.}{Eqs.}
\crefname{figure}{Fig.}{Figs.}
\crefname{section}{Sec.}{Secs.}
\crefname{appendix}{App.}{Apps.}
\crefname{table}{Tab.}{Tabs.}

\newcommand{\sig}{\operatorname{sig}}
\newcommand{\sg}[1]{\mathrm{sg}\!\left(#1\right)}
\DeclareRobustCommand{\RobotAPO}{\texorpdfstring{\textit{RobotAPO}}{RobotAPO}}
\DeclareRobustCommand{\dataset}{\texorpdfstring{\textit{AgiBot-PhysPref}}{AgiBot-PhysPref}}
\newcommand{\graycell}[1]{\textcolor{gray}{#1}}

\renewcommand{\today}{7 October 2026}

\title{RobotAPO: Adversarial Physics Preference Optimization for Robotic Manipulation Video Generation}

\author[1,2]{Kerui Li}
\author[3]{Zhe Jing}
\author[1,2]{Chenyi Huang}
\author[2]{Xiaofeng Wang}
\author[2]{Zheng Zhu}
\author[1,2]{Haoming Cui}
\author[1]{Huaibo Huang}

\affil[1]{Institute of Automation, Chinese Academy of Sciences}
\affil[2]{GigaAI}
\affil[3]{Beijing Institute of Technology}

\hypersetup{
  pdftitle={RobotAPO: Adversarial Physics Preference Optimization for Robotic Manipulation Video Generation},
  pdfauthor={Kerui Li, Zhe Jing, Chenyi Huang, Xiaofeng Wang, Zheng Zhu, Haoming Cui, Huaibo Huang}
}

\begin{document}

\maketitle

\begin{abstract}
Robotic manipulation videos are increasingly used as visual plans for embodied agents, but optimizing purely for visual plausibility often fails to capture the fragile physical manifold of real-world interactions.
Even minor physics-violating errors at the interaction boundary, such as interpenetration or premature object motion, can completely invalidate the inferred timing and pose needed for downstream execution.
Because standard supervised fine-tuning lacks the direct pressure to penalize these localized failures, we introduce \dataset{}.
This rigorously curated $10{,}000$-sample preference dataset isolates condition-matched physics-violations, turning the generator's own failure distribution into a foundational signal for physical consistency.
Building upon this, we propose \RobotAPO{}, an adversarial physics preference optimization framework operating in the continuous flow-matching denoising space.
To prevent the policy from merely memorizing static curated failures, \RobotAPO{} employs a lightweight adversarial counterfactual proposer that learns a condition-dependent, physical-failure-biased direction in denoising space.
This encourages the model to explore and better respect the physical interaction boundary, all while maintaining a pure prompt-and-reference inference interface without requiring external structural conditioning.
Comprehensive evaluations demonstrate that explicitly correcting these localized physics-violations improves downstream robot execution from generated videos.
On held-out AgiBot conditions, \RobotAPO{} outperforms the strongest controlled internal baseline in physical consistency by $6.8\%$ hard score and $10.0\%$ soft score.
Crucially, in real-robot replay, it translates these physical-consistency gains into a $37.4\%$ relative improvement in task success over the strongest controlled internal baseline.
\end{abstract}

\abscontent

\section{Introduction}
\label{sec:intro}

Modern diffusion and image-conditioned video priors have made video synthesis increasingly practical~\citep{latentVideoDiffusion,svd}.
Large text/image-to-video generators further broaden prompt- and reference-conditioned generation~\citep{cogvideo,cogvideox,Vidu}, with recent large-scale video backbones pushing this direction further~\citep{hunyuanvideo,wan}.
Video generation has therefore become a practical interface for embodied intelligence.
Generated videos can serve as task-conditioned visual plans and predictive rollouts for robot learning~\citep{unipi,susie,robodreamer}.
They also support embodied world modeling and policy learning~\citep{vidar,vidman,vpp}.
Related action-learning systems exploit generated or actionless videos as supervision~\citep{avdc,unleashing}.
Synthetic demonstrations and embodied data generation provide another route for scaling robot learning~\citep{uva,dreamgen,gen2act,wow}.
Recent generative transfer pipelines further synthesize robot demonstrations or adapt experience across domains~\citep{gigaworld,robotransfer,embodiedreamer}.

In robotic manipulation, however, visual plausibility is only a partial criterion. 
While supervised fine-tuning (SFT) effectively aligns the model to the broad \textit{visual manifold} of robot appearance and workspace layouts, it often fails to capture the narrow and fragile \textit{physical manifold} governing contact, grasping, and object-state transitions.
A useful generated video must strictly preserve these underlying physical laws. A single-frame physical inconsistency, such as a gripper closing prematurely or an object moving before contact, can instantly invalidate the inferred timing and pose needed for downstream execution.
Because SFT relies on demonstration reconstruction, it receives no direct training pressure to delineate these critical interaction boundaries from harmless perceptual variations.
Consequently, SFT generators continue to produce localized physics-violating errors (e.g., interpenetration, unstable contact) that severely limit their reliability as embodied visual plans.

Preference optimization provides a natural route to contrastive supervision~\citep{rafailov2023dpo,diffusiondpo,zhu2025dspo,liu2025videodpo}.
This motivates a preference signal that is both condition-matched and failure-aware: under the same instruction and first-frame context, a physically grounded manipulation must be strictly preferred over a generated rollout that violates physical causality.
To this end, we introduce \dataset{}, a preference dataset built from $40{,}000$ SFT generator rollouts across $200$ AgiBot task IDs~\citep{agibot}.
Following VLM-based screening and rigorous double human annotation, it provides $10{,}000$ curated pairs isolating local physics-violations.
Because the positive and negative videos share all visual and semantic conditions, this dataset turns the generator's own failure distribution into a foundational signal for physical consistency.

Building upon this insight, we propose \RobotAPO{}, an adversarial physics preference optimization framework in the continuous flow-matching denoising space.
To prevent the policy from merely memorizing static failures, \RobotAPO{} introduces a lightweight \textit{adversarial counterfactual proposer}.
Rather than serving as a standalone discriminator, this proposer learns a condition-dependent, physical-failure-biased direction in the flow-matching denoising space.
It constructs hard negatives that explore the interaction boundary while seeking to preserve the global scene condition.
This mechanism improves preference learning around physics-violating interactions without requiring external structural conditioning (e.g., depth, masks, or physical simulators).
At inference, \RobotAPO{} remains a pure, unconstrained prompt-and-reference-image generator.

We evaluate \RobotAPO{} through automatic video metrics, real-robot replay, and failure-focused qualitative analysis.
Together, these evaluations confirm that correcting localized physics-violations narrows the gap between generated videos and downstream robot execution.
On held-out AgiBot conditions, \RobotAPO{} outperforms the strongest controlled internal baseline in physical consistency.
Crucially, in AnyPos-driven real-robot replay, it translates these physical-consistency gains into a substantial $31.4$ absolute percentage point improvement in task success over the SFT baseline.

Our main contributions are formulated as follows:
\begin{itemize}[leftmargin=1.2em,itemsep=1pt,topsep=2pt]
\item We identify that the critical bottleneck in generative robotic plans lies in physics-violating interaction boundaries. We introduce \dataset{}, a $10{,}000$-sample condition-matched preference dataset explicitly designed to isolate and penalize these local physical failures.
\item We propose \RobotAPO{}, a novel denoising-space preference optimization framework. By introducing an adversarial counterfactual proposer that learns a physical-failure-biased denoising direction, \RobotAPO{} improves physical compliance along the interaction boundary while maintaining a pure prompt-and-reference inference interface.
\item We demonstrate that explicitly penalizing physics violations translates directly to downstream embodied usefulness. Alongside superior automatic physical-consistency metrics, \RobotAPO{} effectively bridges the video-to-action gap, achieving a $64.7\%$ real-robot task success rate with a $37.4\%$ relative improvement over the strongest controlled internal baseline.
\end{itemize}

\section{Related Work}
\label{sec:related_work}

\noindent\textbf{Manipulation video generation.}
Open-domain models provide the generative substrate for robot video synthesis, which are widely used as visual plans~\citep{unipi,susie,robodreamer,anypos}, embodied world models~\citep{vidar,vidman,vpp}, and for demonstration transfer~\citep{gigaworld,robotransfer,emma}.
Most systems rely on reconstruction-based or prediction-based training, effectively capturing the broad visual manifold of scene appearance and coarse dynamics.
However, for embodied execution, visual plausibility is insufficient, and generated plans must strictly respect physical interaction boundaries (e.g., precise contact timing and grasp maintenance).
While recent VLM-based evaluators provide useful diagnostics for general visual and common-sense consistency~\citep{he2024videoscore,he2025videoscore2}, real-world execution remains the ultimate test for these localized physical requirements.

\noindent\textbf{Alignment and preference optimization.}
Preference optimization aligns generative models with criteria beyond standard reconstruction.
Diffusion preference methods implement comparisons in likelihood, score, or denoising spaces~\citep{diffusiondpo,prabhudesai2023aligning_text_to_image,black2024training_diffusion_with_human_feedback,zhu2025dspo}, while recent video frameworks optimize visual quality, semantic alignment, and global plausibility~\citep{liu2025videodpo,realdpo,cheng2025discriminatorfree,hu2025dfusion}.
Concurrently, robot learning heavily exploits self-generated rollouts for data filtering and augmentation~\citep{openxembodiment,openvla,gr2,dreamgen}.
Despite these advances, existing objectives rarely target the highly localized, action-critical physics-violations inherent in robotic manipulation. 
To bridge this gap, our work explicitly aligns the generation process with physical causality at the interaction boundary rather than generic visual preferences.

\section{Method}
\label{sec:method}

\subsection{Problem and Overview}
\label{sec:method_overview}

Throughout the paper, backbone refers only to the Wan2.2-TI2V architecture.
We refer to the robot-data-specialized checkpoint before adding the \RobotAPO{} LoRA adapters as the SFT generator.
This same checkpoint supplies the frozen SFT-centering scores during optimization and serves as the SFT baseline in experiments.
While the SFT generator captures the broad visual manifold of robotic workspaces, its rollouts frequently deviate from the underlying physical manifold.
We post-train this generator to explicitly respect physical interaction boundaries, ensuring its samples are both visually plausible and useful for execution-oriented visual planning.

A manipulation condition $c$ contains a task instruction and a first-frame visual context.
Each training example compares two videos under the same condition.
One is a real successful demonstration that lies safely within the physical manifold.
The other is a generated rollout that falls off this manifold, exhibiting a localized physics-violating error.
We write each condition-matched preference sample as $(c,\,x^{+},\,x^{-},\,e^{-})$ with $x^{+}\succ x^{-}$.
Here, $x^{+}$ is the corresponding real AgiBot demonstration under $c$.
$x^{-}$ is a rollout from the SFT generator that exhibits a visible local physical failure under the same $c$.
$e^{-}$ is the primary failure label, chosen as the most severe and visually salient error when multiple failures are present.
Because $x^{+}$ and $x^{-}$ share instruction, scene, embodiment, and visual condition, the binary preference $x^{+}\succ x^{-}$ reduces scene- and semantic-level variation and focuses supervision on physical interaction validity. The remaining real-versus-generated source asymmetry is examined with a source-matched control in \cref{sec:add_gg}.

\begin{figure*}[t]
\centering
\includegraphics[width=\textwidth]{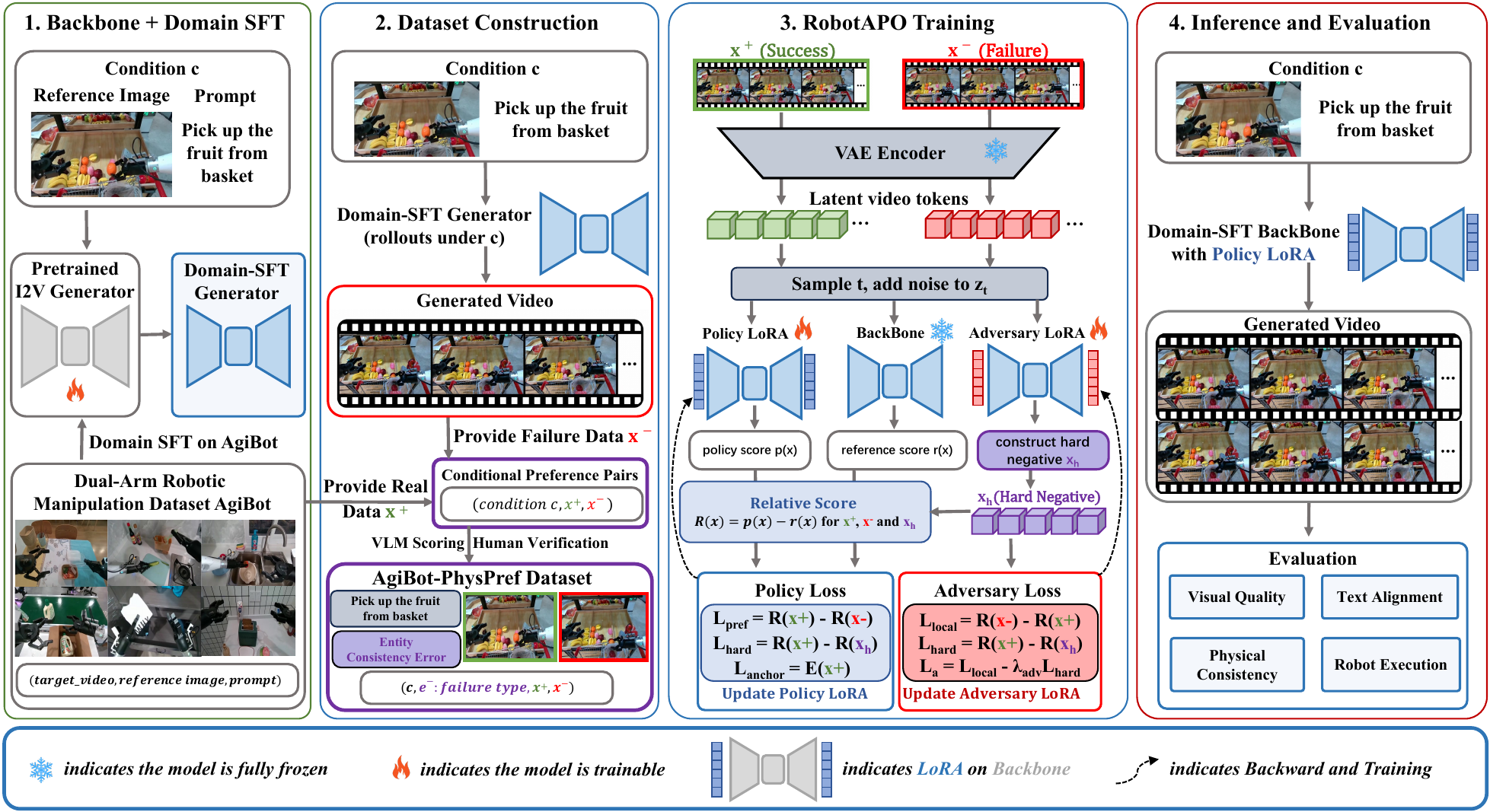}
\vspace{-10pt}
\caption{
Overview of \RobotAPO{}. 
SFT rollouts are screened and annotated to build the condition-matched \dataset{}. 
During post-training, an adversarial proposer learns a physical-failure-biased denoising direction to probe interaction boundaries, encouraging the policy to better respect the physical manifold.
At inference, only the policy generator is deployed, maintaining a standard prompt-and-reference interface without any external structural conditioning.
Displayed scores are schematic, while exact denoising rewards and objectives are defined in Sec.~\ref{sec:method_objective}.
}
\label{fig:pipeline}
\vspace{-10pt}
\end{figure*}

\RobotAPO{} (\cref{fig:pipeline}) attaches two low-rank adaptation (LoRA) modules to the frozen Wan2.2 text-image-to-video (TI2V) SFT generator~\citep{wan}: a trainable policy generator $f_{\pi}$ and a training-only adversarial proposer $f_{a}$.
During optimization, the frozen SFT generator serves as a stop-gradient centering model for DPO-style scoring. 
While preference pairs penalize known condition-matched failures, $f_{a}$ explores nearby invalid interactions omitted by the static dataset. 
This encourages $f_{\pi}$ to optimize against condition-preserving negatives. 
At inference, all auxiliary branches are discarded and only $f_{\pi}$ is deployed. 
It retains the standard prompt-and-reference inference interface, requiring no external structural conditioning (e.g., action, trajectory, masks, or depth).

\subsection{AgiBot-PhysPref Construction}
\label{sec:method_pairs}

\textbf{Failure taxonomy.}
We adopt a unified five-category taxonomy of visible local physical failures: interpenetration, entity consistency, contact causality, unsupported floating after release, and invalid interaction/other.
Interpenetration covers robot, object, or scene geometry passing through another entity.
Entity consistency covers changes in object appearance, shape, identity, or existence over time.
Contact causality covers object motion or state transitions without valid contact or manipulation, including premature motion before effective contact.
Unsupported floating covers objects that remain suspended after losing support, while invalid interaction/other covers missed contacts, grasping air, or task-condition violations.
The same taxonomy is used for human annotation, qualitative analysis, and the failure-focused figure. Full definitions are provided in \cref{sec:supp_failure_taxonomy}.

\textbf{Construction pipeline.}
Starting from manipulation conditions in the AgiBot robot-manipulation benchmark~\citep{agibot}, we roll out the SFT generator to obtain a candidate pool of $40{,}000$ generated manipulation videos covering $200$ task IDs over dual-arm scenarios.
We then use Qwen3-VL-30B-Instruct as a quality-control screener.
It ranks candidates by robot-interaction quality, 3D consistency, and instruction following.
Annotators inspect candidates in descending screening order until $10{,}000$ negatives are accepted. They review $12{,}935$ candidates and exclude $2{,}935$ videos that lack a visible physical failure or sufficient evidence for reliable labeling.
Every reviewed candidate is independently assessed as a full video by two mutually blinded annotators, with disagreements sent to panel adjudication. The pre-adjudication seven-way judgments agree on $10{,}579/12{,}935$ candidates ($81.79\%$), with nominal Krippendorff's $\alpha=0.7832$. Every sample in the final set has a human-verified physical-failure label.
When multiple visible failures are present, annotators record the primary label as the most severe and visually salient error.
Each verified failed rollout is paired with the matching real AgiBot demonstration under the same condition, producing the curated $10{,}000$-sample condition-matched preference set.
The Qwen3-VL screening prompt, annotation protocol, and failure taxonomy are provided in
\cref{sec:supp_dataset_artifacts,sec:supp_dataset_flow,sec:supp_vlm_prompt,sec:supp_human_protocol,sec:supp_failure_taxonomy}.

We denote the resulting preference set by $\mathcal{D}_{\mathrm{phy}}=\{(c_i,\, x_i^{+},\, x_i^{-},\, e_i^{-})\}_{i=1}^{N}$ with $N=10{,}000$, on which training only uses the binary relation $x_i^{+} \succ x_i^{-}$.
Holding the condition fixed prevents the model from picking up spurious scene-level preferences and forces optimization to focus on grasp, contact, and object-motion quality.
The primary failure annotations $\{e_i^{-}\}$ are kept for balancing the training mix across the five categories and for failure-focused qualitative analysis.
They are metadata rather than multi-class rewards.
Overlapping visible failures can be noted during review, but training stores the primary label.

\begin{table}[t]
\centering
\caption{
\dataset{} dataset statistics. 
The curated preference set isolates local physics violations via condition-matched pairs, human verification, and Qwen3-VL screening.
}
\label{tab:dataset_stats}
\small
\begin{tabularx}{\linewidth}{lX}
\toprule
\textbf{Property} & \textbf{Value} \\
\midrule
Generated candidates & $40{,}000$ \\
Human-reviewed candidates & $12{,}935$ \\
Final curated preference samples & $10{,}000$ \\
Task IDs & $200$ dual-arm AgiBot scenarios \\
Screening model & Qwen3-VL-30B-Instruct \\
Final label source & Human double annotation with disagreement review \\
Pre-adjudication agreement & $81.79\%$; Krippendorff's $\alpha=0.7832$ \\
Preference unit & Condition-matched tuple $(c,\,x^{+},\,x^{-},\,e^{-})$ \\
Failure focus & Human-verified visible local physical failures \\
Failure labels & Interpenetration, entity consistency, contact causality, unsupported floating, invalid interaction/other \\
\bottomrule
\end{tabularx}
\end{table}

\subsection{Denoising-Space Physical-Failure Preference Learning}
\label{sec:method_objective}

\RobotAPO{} instantiates pairwise preference learning in the same flow-matching denoising space used to train the video generator.
For a video $x$, let $z_0$ be its variational autoencoder (VAE) latent.
For each training pair, we sample a flow-matching noise level $\sigma$ and Gaussian noise $\epsilon$.
Under the rectified-flow schedule used in our Wan2.2 implementation, $z_{\sigma}(x)=(1-\sigma)z_0(x)+\sigma\epsilon$ and $u_{\sigma}(x)=\epsilon-z_0(x)$, so a clean-latent estimate from a velocity prediction $\hat{u}$ is $z_{\sigma}-\sigma\hat{u}$.
The task instruction and first-frame reference image are fed through the same image-to-video conditioning mechanism as the SFT generator. We denote the resulting condition by $c$.
We package the denoising target as $y_x=(z_{\sigma}(x),u_{\sigma}(x))$ for a video $x$.
Thus the preferred demonstration and curated failed rollout produce $y_{x^{+}}$ and $y_{x^{-}}$, respectively.
Counterfactual hard negatives introduced later use the same target format.

For a generator $b\in\{\pi,a,\mathrm{SFT}\}$ with velocity prediction $\hat{u}_b(z,\sigma,c)$ and any denoising target $y=(z_y,u_y)$, define the denoising energy as
\begin{equation}
\mathcal{E}_b(y,\sigma,c)=\bigl\|\hat{u}_{b}(z_y,\sigma,c)-u_y\bigr\|_2^2.
\label{eq:denoise_energy}
\end{equation}
In implementation this squared error is averaged over latent dimensions and minibatch samples.
For readability in the loss formulas, we write $\mathcal{E}_b(y)$ as shorthand for $\mathcal{E}_b(y,\sigma,c)$ when $\sigma$ and $c$ are fixed within a training pair.
The operator $\sg{\cdot}$ denotes stop-gradient: its argument is used as a constant and receives no gradient.
For either trainable branch $b\in\{\pi,a\}$, we define the same SFT-centered denoising reward
\begin{equation}
R_b(y)=-\mathcal{E}_b(y)+\sg{\mathcal{E}_{\mathrm{SFT}}(y)}.
\label{eq:denoise_reward}
\end{equation}
A larger $R_b(y)$ means that branch $b$ assigns lower denoising error to $y$ relative to the frozen SFT generator.
In other words, $R_{\pi}$ and $R_a$ are the policy and adversary instances of the same score, with the same formula and different branch weights.
The losses below compare reward differences between successful and negative targets, which keeps the DPO-style comparison local to the starting checkpoint while making the manipulation-specific negative targets the source of preference signal.

\textbf{Curated physical-failure preference.}
We write $\sig(\cdot)$ for the logistic sigmoid, kept distinct from the flow-matching noise level $\sigma$.
The scalar $\gamma>0$ is the preference temperature that scales the reward difference before the sigmoid.
For a condition-matched physical-failure pair $(c, x^{+}, x^{-})$, the real successful video contributes $y_{x^{+}}$ and the curated failed rollout contributes $y_{x^{-}}$.
This keeps the pairwise preference form of DPO~\citep{rafailov2023dpo,diffusiondpo,zhu2025dspo}, but the comparison is explicitly grounded in flow-matching denoising energy and condition-matched manipulation failures.
Both targets are evaluated under the same prompt, first-frame reference, noise level, and target construction, so the preference signal is aimed at local interaction errors rather than semantic shifts.

\textbf{Preferred-sample anchor.}
To prevent the pairwise update from over-amplifying and damaging valid behavior of the SFT generator, we use a preferred-sample anchor and a LoRA drift penalty:
\begin{equation}
\mathcal{L}_{\mathrm{anc}}=\mathcal{E}_{\pi}(y_{x^{+}},\sigma,c),\qquad
\mathcal{L}_{\mathrm{reg}}=\bigl\|\mathrm{LoRA}(f_{\pi})\bigr\|_2^2.
\label{eq:l_anchor_reg}
\end{equation}
The anchor keeps the policy close to valid denoising behavior on successful trajectories, and the regularizer is a small weight penalty on the policy LoRA adapter.

\subsection{Adversarial Counterfactual Hard-Negative Generation}
\label{sec:method_adv}

Static curated failures become less informative once the policy learns to avoid obvious visual artifacts.
To keep the negatives informative, \RobotAPO{} introduces an adversarial counterfactual mechanism that constructs hard negatives near the physical interaction boundary.

\textbf{Hard-negative construction.}
Given a successful video $x^{+}$ and its noisy latent $z_{\sigma}^{+}$, we construct an adversarial counterfactual target by blending the policy and adversarial velocity predictions with a small mixing weight $\lambda$ sampled from the range reported in \cref{sec:supp_implementation}:
\begin{equation}
\hat{u}_{h}=\hat{u}_{\pi}(z_{\sigma}^{+},\sigma,c)+\lambda\!\left[\hat{u}_{a}(z_{\sigma}^{+},\sigma,c)-\hat{u}_{\pi}(z_{\sigma}^{+},\sigma,c)\right].
\label{eq:hard_pred_blend}
\end{equation}
Here $\lambda$ is only the hard-negative blending coefficient in latent denoising space. It is distinct from the adversary loss weight $\lambda_{\mathrm{adv}}$ introduced below.
The counterfactual clean-latent estimate is $z_0^{h}=z_{\sigma}^{+}-\sigma\hat{u}_{h}$.
We write the hard-negative target as
\begin{equation}
y_h=h_a(y_{x^{+}},\sigma,c,\lambda).
\label{eq:hard_negative_target}
\end{equation}
Concretely, $y_h=(z_{\sigma}^{+},u_h)$ with $u_h\equiv\hat{u}_h$, whose implied clean latent under the same flow schedule is $z_0^h$.
Because this perturbation starts from the successful noisy latent $z_{\sigma}^{+}$, it is designed to preserve the condition while moving toward interactions that reduce the policy's preference margin.
The decoded comparison in \cref{sec:add_gaussian} evaluates whether the learned direction preferentially exposes local physical failures rather than generic corruption.

\textbf{RobotAPO preference loss and policy objective.}
The preference term ranks the successful target above two manipulation-specific negatives: the curated failed rollout and the adversarial counterfactual target.
We denote the negative target set for the current training example as
\begin{equation}
\mathcal{N}_{\mathrm{neg}}=\{\,y_{x^{-}},\,y_h\,\}.
\label{eq:negative_set}
\end{equation}
\begin{equation}
\mathcal{L}_{\mathrm{pref}}
=\sum_{y^{-}\in\mathcal{N}_{\mathrm{neg}}}
-\log \sig\!\Bigl(\gamma\bigl(R_{\pi}(y_{x^{+}})-R_{\pi}(y^{-})\bigr)\Bigr).
\label{eq:l_pref}
\end{equation}
The first summand uses the curated failed rollout $y_{x^{-}}$, and the second summand uses the counterfactual hard negative $y_h$.
The full policy objective is then
\begin{equation}
\mathcal{L}_{\pi}=\mathcal{L}_{\mathrm{pref}}+\mathcal{L}_{\mathrm{anc}}+\mathcal{L}_{\mathrm{reg}},
\label{eq:l_policy}
\end{equation}
Here $\mathcal{L}_{\mathrm{anc}}$ and $\mathcal{L}_{\mathrm{reg}}$ are defined in \cref{eq:l_anchor_reg}.
During a policy update, only the LoRA parameters of $f_{\pi}$ are updated, while $f_a$ and the SFT centering model are frozen.

\textbf{Adversary objective.}
The adversarial proposer is a generator-side module rather than a discriminator.
It does not score decoded videos directly and is never decoded to pixels at inference.
Instead, it combines an inverted pairwise signal on the real success-vs-failure pair with a hard-negative pressure. 
During this update, the policy weights are frozen, but gradients still flow back through the adversary-induced hard target $y_h$:
\begin{equation}
\mathcal{L}_{\mathrm{hard}}^{(a)}
=-\log \sig\!\Bigl(\gamma\bigl(R_{\pi}(y_{x^{+}})-R_{\pi}(y_h)\bigr)\Bigr)
\label{eq:l_cf_adv}
\end{equation}
The policy term is evaluated with stop-gradient parameters, and the counterfactual target $y_h$ remains differentiable with respect to $f_a$ through \cref{eq:hard_pred_blend,eq:hard_negative_target}.
The adversary-side local preference and full objective are
\begin{equation}
\mathcal{L}_{\mathrm{local}}^{(a)}=-\log \sig\!\Bigl(\gamma\bigl(R_{a}(y_{x^{-}})-R_{a}(y_{x^{+}})\bigr)\Bigr)
\label{eq:l_local_adv}
\end{equation}
\begin{equation}
\mathcal{L}_{a}
=\mathcal{L}_{\mathrm{local}}^{(a)}-\lambda_{\mathrm{adv}}\mathcal{L}_{\mathrm{hard}}^{(a)}
\label{eq:l_adv}
\end{equation}
The first term uses the same reward form as \cref{eq:denoise_reward}, evaluated on the adversary branch, and makes the proposer prefer the failed target over the successful one.
The second term increases the policy's counterfactual loss, with $\lambda_{\mathrm{adv}}$ controlling the coupling strength.

Equations~\ref{eq:l_policy} and \ref{eq:l_adv} establish a dynamic Min-Max game around the physical interaction boundary.
The adversary searches for plausible velocity perturbations that reduce the policy's preference margin, while the policy is trained to increase this margin.
This boundary exploration is performed in the continuous denoising space and does not require external 3D physical simulators, depth maps, or mask constraints.
Training begins with a policy-only warm-up and then alternates between policy and adversary phases, updating only the active branch while the SFT centering model remains frozen throughout.
The $\lambda$ sampling range, optimizer settings, and full schedule are reported in \cref{sec:supp_implementation}.

\section{Experiments}
\label{sec:exp}

We design our experiments to validate whether explicitly optimizing for physical causality improves downstream robot execution from generated videos.
We evaluate generated videos along three axes:
1) Automatic visual, alignment, and physical-consistency scoring,
2) Downstream embodied usefulness through real-robot pick-and-place execution, and
3) Local physical behavior through failure-focused qualitative analysis at the interaction boundary.

\subsection{Experimental Setup}
\label{sec:exp_setup}

\noindent\textbf{Backbone, data, and fair comparison.}
To strictly isolate the impact of training objectives, all internal methods share the Wan2.2-TI2V backbone~\citep{wan}, SFT initialization, condition-matched preference pairs, and inference settings (prompts, reference images, and samplers).
The SFT baseline is evaluated without LoRA adapters, while Diffusion-DPO~\citep{diffusiondpo} and RealDPO~\citep{realdpo} serve as generic preference baselines.
The full \RobotAPO{} integrates anchor-preserving regularization and adversarial counterfactuals. We ablate these mechanisms via the \textit{w/o Anchor/Reg} and \textit{w/o CF/Adv} variants, respectively.
To contextualize our 5B framework's absolute performance, we also include LVP~\citep{chen2025lvp} (Wan2.1-I2V, 14B) as a high-capacity external scaling reference. 
Evaluation protocols for automatic scoring, robot replay, and qualitative inspection are detailed in their respective subsections.

\subsection{Main Quantitative Comparison}
\label{sec:exp_main}

For the AgiBot automatic comparison, we first train the SFT generator on curated dual-arm manipulation videos~\citep{agibot}.
All internal preference methods initialize from it, train on \dataset{}, and are evaluated on disjoint held-out conditions with identical prompts, reference images, sampler settings, a fixed shared LoRA scale of $0.4$, and random seeds.
We use the official VideoScore2 evaluator~\citep{he2025videoscore2} on $200$ held-out task conditions with five samples per condition, for $1{,}000$ videos per method.
We report three VideoScore2~\citep{he2025videoscore2} dimensions: visual quality, text-to-video alignment, and physical/common-sense consistency.
For each dimension, \textbf{H} denotes the parsed hard integer score and \textbf{S} denotes the logit-based soft score, both on a $1$--$5$ scale.

\begin{table}[t]
\centering
\caption{
Quantitative comparison on held-out AgiBot conditions. 
All internal methods are controlled 5B-parameter models. 
\RobotAPO{} outperforms all controlled internal baselines; the external 14B LVP reference is shown in gray to contextualize absolute performance across a different model and training setup.
Bold marks the best internal method.
}
\label{tab:main_results}
\begingroup
\small
\setlength{\tabcolsep}{0pt}
\begin{tabular*}{\linewidth}{@{\extracolsep{\fill}}llcccccc@{}}
\toprule
Method
& Size
& Visual-H$\uparrow$
& Align-H$\uparrow$
& Phys-H$\uparrow$
& Visual-S$\uparrow$
& Align-S$\uparrow$
& Phys-S$\uparrow$ \\
\midrule
\graycell{LVP~\citep{chen2025lvp}}
& \graycell{14B} & \graycell{3.647} & \graycell{3.787} & \graycell{3.473} & \graycell{3.030} & \graycell{3.252} & \graycell{3.363} \\
\midrule
Wan2.2-TI2V
& 5B & 3.287 & 3.493 & 2.967 & 2.812 & 3.064 & 2.891 \\
SFT baseline
& 5B & 3.347 & 3.767 & 3.193 & 2.984 & 3.410 & 3.041 \\
Diffusion-DPO~\citep{diffusiondpo}
& 5B & 3.340 & 3.760 & 3.253 & 3.120 & 3.484 & 3.172 \\
RealDPO~\citep{realdpo}
& 5B & 3.347 & 3.760 & 3.173 & 3.047 & 3.435 & 3.008 \\
\midrule
{\RobotAPO{} (Ours)}
& 5B & \textbf{3.553} & \textbf{3.927} & \textbf{3.473} & \textbf{3.564} & \textbf{3.939} & \textbf{3.489} \\
\bottomrule
\end{tabular*}
\endgroup
\end{table}

\Cref{tab:main_results} reports official VideoScore2 scores on the held-out evaluation split.
The gap between Wan2.2-TI2V and the SFT baseline confirms that robot-domain supervised fine-tuning is necessary to adapt the foundation prior to manipulation scenes.
Compared with the SFT baseline, \RobotAPO{} raises the hard scores for visual quality, alignment, and physical consistency to $3.553$, $3.927$, and $3.473$, respectively.
\RobotAPO{} also improves the soft scores by $0.580$ for visual quality, $0.529$ for alignment, and $0.448$ for physical consistency.

Among all internal Wan2.2-TI2V methods, \RobotAPO{} sets the best mark on every hard and soft metric.
Diffusion-DPO improves physical consistency over the SFT baseline but leaves the hard visual and alignment scores nearly unchanged.
RealDPO yields smaller and less consistent gains, with physical scores occasionally falling below the SFT baseline.
Together, these controlled comparisons suggest that generic preference objectives struggle to isolate the fragile physical manifold. The failure-aware supervision and adversarial hard-negative design of \RobotAPO{} are essential for this setting.
Crucially, visual quality also improves concurrently, indicating that strictly enforcing physical interaction boundaries does not trade off against global video plausibility.

The 14B external baseline, LVP, attains the highest hard visual-quality score and ties \RobotAPO{} on Phys-H, reflecting the representational strength of its larger parameter scale.
At roughly one-third the parameter scale, \RobotAPO{} has the same observed Phys-H and higher observed soft scores and Align-H in this evaluation. Since LVP is an external reference rather than a matched-training baseline, these values contextualize performance but do not isolate the effect of the RobotAPO objective.

\subsection{Robot Execution from Generated Videos}
\label{sec:exp_robot}

While automatic metrics capture perceptual plausibility, the ultimate test of physical manifold adherence is downstream execution.
We run a separate real-robot replay evaluation on an AgileX platform and video domain.
SFT, Diffusion-DPO, RealDPO, and \RobotAPO{} start from their respective AgiBot-trained checkpoints and undergo target-domain adaptation on the same $400$ self-collected AgileX videos, with matched optimization schedules and trainable-parameter budgets.
Thus, the held-out AgiBot experiment evaluates \dataset{} directly, while the AgileX study evaluates limited-data cross-platform adaptation from the corresponding AgiBot-trained checkpoints.
LVP is evaluated directly from its released 14B checkpoint to provide a high-capacity, out-of-the-box reference point.
All generated videos are passed through the same frozen AnyPos video-to-action pipeline~\citep{anypos} and controller.
The recovered actions are replayed on $51$ matched trials ($17$ task-scene configurations, three repeats each).
A total of 51 videos per method and 255 videos overall are retained and executed without filtering, regeneration, trajectory correction, or manual intervention; every video yields a finite controller-accepted trajectory.
A trial is successful only if the robot completes the full pick-and-place instruction.
We report a Wilson $95\%$ confidence interval for task success.

\begin{table}[t]
\centering
\caption{
Real-robot pick-and-place execution success rates via AnyPos action extraction. 
\RobotAPO{} improves downstream robot execution from generated videos, raising absolute task success by $31.4$ percentage points over the SFT baseline. LVP is an external, unmatched reference.
(Task success 95\% CI uses the Wilson score interval over $51$ trials).
}
\label{tab:robot_execution}
\begingroup
\small
\setlength{\tabcolsep}{0pt}
\begin{tabular*}{\linewidth}{@{\extracolsep{\fill}}llcccc@{}}
\toprule
Method
& Size
& \begin{tabular}[c]{@{}c@{}}Grasp\\Success (\%)$\uparrow$\end{tabular}
& \begin{tabular}[c]{@{}c@{}}Cond. Place\\Success (\%)$\uparrow$\end{tabular}
& \begin{tabular}[c]{@{}c@{}}Task\\Success (\%)$\uparrow$\end{tabular}
& \begin{tabular}[c]{@{}c@{}}Task Success\\95\% CI\end{tabular} \\
\midrule
\graycell{LVP} & \graycell{14B} & \graycell{72.5} & \graycell{86.5} & \graycell{62.7} & \graycell{[49.0, 74.7]} \\
\midrule
SFT baseline
& 5B & 56.9 & 58.6 & 33.3 & [22.0, 47.0] \\
Diffusion-DPO
& 5B & 64.7 & 69.7 & 45.1 & [32.3, 58.6] \\
RealDPO
& 5B & 66.7 & 70.6 & 47.1 & [34.1, 60.5] \\
\RobotAPO{} (Ours)
& 5B & \textbf{78.4} & \textbf{82.5} & \textbf{64.7} & [51.0, 76.4] \\
\bottomrule
\end{tabular*}
\endgroup
\end{table}

\Cref{tab:robot_execution} shows that \RobotAPO{} attains the highest observed task success rate ($64.7\%$, $33/51$).
It improves over the SFT baseline by $31.4$ absolute percentage points under the matched adaptation and AnyPos extraction pipeline; the configuration-clustered paired 95\% CI is $[15.7,49.0]$ percentage points.
The 14B external baseline, LVP, reaches $62.7\%$ observed task success. The paired RobotAPO--LVP difference is $+2.0$ percentage points with a 95\% CI of $[-9.8,13.7]$, so the two systems attain similar observed success under the current trial budget rather than establishing a significant advantage over LVP. These results evaluate the end-to-end combination of each generator with the frozen AnyPos/controller interface.

\subsection{Ablation Study}
\label{sec:exp_ablation}

\begin{table}[t]
\centering
\caption{
Module ablation of \RobotAPO{}. 
Removing the adversarial counterfactual proposer (w/o CF/Adv) reduces real-robot task success from $64.7\%$ to $49.0\%$, supporting the contribution of dynamic boundary exploration in this setting.
}
\label{tab:module_ablation}
\begingroup
\small
\setlength{\tabcolsep}{0pt}
\begin{tabular*}{\linewidth}{@{\extracolsep{\fill}}lccccc@{}}
\toprule
Method
& Visual-S$\uparrow$
& Align-S$\uparrow$
& Phys-S$\uparrow$
& \begin{tabular}[c]{@{}c@{}}Grasp\\Success (\%)$\uparrow$\end{tabular}
& \begin{tabular}[c]{@{}c@{}}Task\\Success (\%)$\uparrow$\end{tabular} \\
\midrule
SFT baseline
& 2.984 & 3.410 & 3.041 & 56.9 & 33.3 \\
\RobotAPO{} w/o CF/Adv
& 3.194 & 3.614 & 3.134 & 68.6 & 49.0 \\
\RobotAPO{} w/o Anchor/Reg
& 3.000 & 3.326 & 3.079 & 60.8 & 35.3 \\
{\RobotAPO{} (Ours)}
& \textbf{3.564} & \textbf{3.939} & \textbf{3.489} & \textbf{78.4} & \textbf{64.7} \\
\bottomrule
\end{tabular*}
\endgroup
\end{table}

\noindent\textbf{Module ablation.}
We ablate the two main design axes introduced by \RobotAPO{}: the counterfactual adversarial hard-negative mechanism (w/o CF/Adv) and the preferred-sample anchor with regularization (w/o Anchor/Reg).
\Cref{tab:module_ablation} provides the most critical insight into the system's performance.
Static curated failures alone (w/o CF/Adv) are sufficient to improve generic VideoScore2 metrics (Visual-S rises to $3.194$, Align-S to $3.614$).
However, without the dynamic adversarial boundary exploration, Task Success drops precipitously by $15.7$ absolute percentage points (from $64.7\%$ to $49.0\%$).
This contrast supports our core hypothesis: while static DPO can align models to broad visual and semantic preferences, the learned physical-failure-biased direction provides additional pressure around interaction failures that matter for downstream execution.
Furthermore, removing the anchor and regularization (w/o Anchor/Reg) collapses both alignment and physical-consistency scores, validating their role in preventing the policy from severely deviating from the valid denoising manifold established during SFT.

\noindent\textbf{Sensitivity to $\lambda_{\mathrm{adv}}$.}
We examine the adversarial coupling coefficient $\lambda_{\mathrm{adv}}$ in \cref{tab:lambda_adv_sensitivity}, which controls the intensity of the counterfactual physical boundary perturbation.
A weak adversarial coupling ($\lambda_{\mathrm{adv}}=0.02$) fails to provide sufficient pressure to explore the interaction boundary, yielding marginal gains.
Conversely, an overly aggressive coupling ($\lambda_{\mathrm{adv}}=0.10$) disrupts the underlying visual manifold, harming both alignment and physical scores.
The default setting ($0.05$) best balances aggressive adversarial boundary exploration with the structural stability of the visual manifold.

\begin{table}[t]
\centering
\vspace{-15pt}
\caption{
Sensitivity study of the adversarial coupling coefficient $\lambda_{\mathrm{adv}}$.
This coefficient controls the strength of the adversarial counterfactual term that perturbs the local physical interaction boundary during training.
The dagger $\dagger$ marks the default setting used in \RobotAPO{}.
}
\label{tab:lambda_adv_sensitivity}
\begingroup
\small
\setlength{\tabcolsep}{0pt}
\begin{tabular*}{\linewidth}{@{\extracolsep{\fill}}lcccccc@{}}
\toprule
$\lambda_{\mathrm{adv}}$
& Visual-H$\uparrow$
& Align-H$\uparrow$
& Phys-H$\uparrow$
& Visual-S$\uparrow$
& Align-S$\uparrow$
& Phys-S$\uparrow$ \\
\midrule
$0.00$
& 3.407 & 3.793 & 3.300 & 3.284 & 3.704 & 3.270 \\
$0.02$
& 3.460 & 3.853 & 3.373 & 3.392 & 3.784 & 3.351 \\
$0.05^\dagger$
& \textbf{3.553} & \textbf{3.927} & \textbf{3.473} & \textbf{3.564} & \textbf{3.939} & \textbf{3.489} \\
$0.10$
& 3.433 & 3.827 & 3.420 & 3.365 & 3.758 & 3.421 \\
\bottomrule
\end{tabular*}
\endgroup
\end{table}

\subsection{Failure-Focused Qualitative Comparison}
\label{sec:exp_failure_qual}

\begin{figure}[t]
\centering
\IfFileExists{figures/Q3.pdf}{
\includegraphics[width=\linewidth]{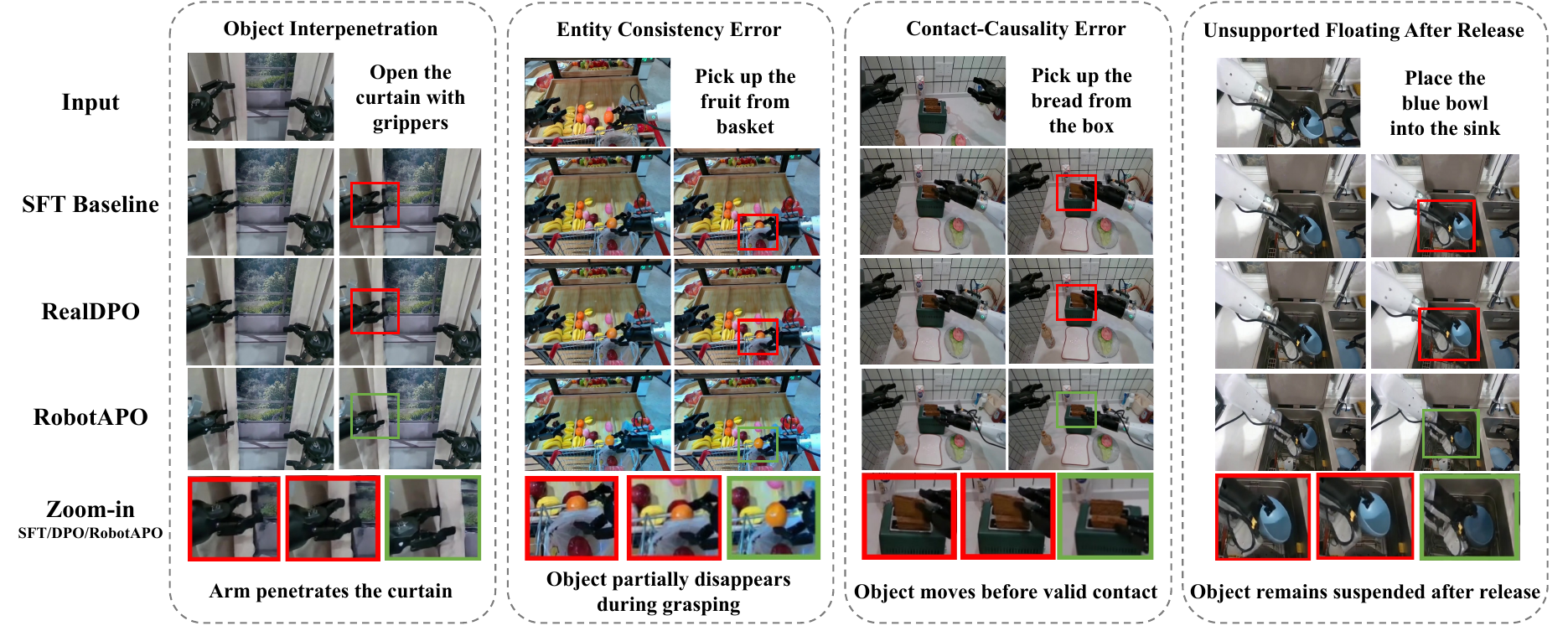}
}{
\fbox{\parbox{0.92\linewidth}{\centering
Placeholder for failure-focused qualitative comparison.
Recommended layout: each column shows one representative failure type from the unified five-category taxonomy.
The input row gives the reference image and instruction.
Method rows show two key frames per method (before / at the local error).
The bottom row enlarges the highlighted regions.}}
}
\vspace{-15pt}
\caption{
Failure-focused qualitative comparison on held-out conditions. 
For each task condition, all methods generate videos with the same frame length and inference settings. We visualize frames aligned to the same action-critical phase.
Red boxes highlight physics-violating phenomena (e.g., interpenetration, premature object motion) frequently produced by baselines at action-critical moments. 
Green boxes show that \RobotAPO{} reduces these local interaction artifacts and better preserves gripper--object geometry and contact causality.
Zoom-ins isolate the key physical interactions.
}
\label{fig:qualitative_failure}
\vspace{-12pt}
\end{figure}

\Cref{fig:qualitative_failure} compares generated rollouts organized by our five-category failure taxonomy.
Rather than presenting generic success cases, this analysis isolates critical interaction boundaries.
While baselines may maintain high overall visual quality, they frequently show physical-causality errors at action-critical moments (e.g., interpenetrating the curtain, or the object shifting prematurely before the gripper closes).
\RobotAPO{} reduces these local interaction artifacts and better preserves gripper--object geometry, contact timing, and unsupported post-release physics.
These qualitative boundary corrections are consistent with the gains in real-robot grasp and task success rates.

\section{Conclusion}
\label{sec:conclusion}

We presented \RobotAPO{}, an adversarial preference optimization framework for improving physical consistency in robotic video generation.
Leveraging our condition-matched \dataset{}, \RobotAPO{} optimizes physical preferences directly within the continuous flow-matching denoising space.
To avoid memorizing static artifacts, an adversarial proposer learns a condition-dependent, physical-failure-biased denoising direction.
This endogenous boundary exploration improves behavior around physical interaction failures without relying on external structural constraints.
Real-robot evaluations show that reducing these localized physics violations improves end-to-end video-to-action execution under a fixed action-extraction interface.
Future work will scale this paradigm and integrate closed-loop execution signals to further align generative visual plans with reliable physical control.

\bibliographystyle{plainnat}
\bibliography{main}

\clearpage
\appendix
\section{Additional Controlled Analyses}
\label{sec:add_controls}

\subsection{Source-Matched Generated-vs.-Generated Preferences}
\label{sec:add_gg}

The original \dataset{} construction pairs a real successful demonstration with a generated rollout containing a human-verified physical failure. Although the instruction and first-frame condition are matched, this construction does not remove the real-versus-generated source difference. We therefore conduct a source-matched generated-vs.-generated (G+G) control in which both videos in each preference pair are sampled from the same Wan2.2-TI2V SFT generator under the same AgiBot training condition.

For each condition, we generate ten rollouts with different seeds and rank them using a confidence-aware binary physical-error classifier trained only on \dataset{} human labels, without VideoScore2 supervision. The least-flawed rollout is preferred and the most-flawed rollout is rejected. We construct $12{,}935$ candidate pairs and remove the $2{,}935$ pairs with the smallest classifier-margin gaps, retaining $10{,}000$ G+G pairs. Independent human checking on 100 randomly sampled pairs yields an aggregate agreement of $84.5\%$ with the classifier ordering.

Diffusion-DPO, RealDPO, and \RobotAPO{} are retrained from the same SFT initialization with identical G+G data and training budgets. Evaluation follows the original protocol of 200 held-out conditions, five generations per condition, and $1{,}000$ videos per method.

\begin{table}[t]
\centering
\caption{Source-matched G+G control. Both sides of every training preference pair are generated by the same SFT model under the same condition.}
\label{tab:add_gg}
\small
\begin{tabular}{lcccc}
\toprule
Method & Visual-H$\uparrow$ & Visual-S$\uparrow$ & Phys-H$\uparrow$ & Phys-S$\uparrow$ \\
\midrule
SFT & 3.347 & 2.984 & 3.193 & 3.041 \\
Diffusion-DPO & 3.240 & 3.211 & 3.185 & 3.061 \\
RealDPO & 3.225 & 3.212 & 3.180 & 3.092 \\
\RobotAPO{} (Ours) & \textbf{3.355} & \textbf{3.345} & \textbf{3.310} & \textbf{3.293} \\
\bottomrule
\end{tabular}
\end{table}

Under source-matched supervision, \RobotAPO{} improves Phys-S over Diffusion-DPO by $+0.232$ (95\% CI $[0.132,0.336]$) and over RealDPO by $+0.201$ ($[0.120,0.289]$). The corresponding Phys-H gains are $+0.125$ ($[0.060,0.190]$) and $+0.130$ ($[0.070,0.190]$). These condition-level paired bootstrap intervals resample the 200 conditions while retaining all five generations within each sampled condition. The persistent gains show that the advantage of \RobotAPO{} cannot be explained solely by a direct source-identity cue.

\subsection{Learned Direction vs. Matched Gaussian Perturbations}
\label{sec:add_gaussian}

We test whether the adversarial residual captures failure-related structure beyond generic latent corruption. With the checkpoint and hyperparameters frozen, we randomly select 64 held-out conditions before decoding and generate one learned-adversarial and one equal-norm Gaussian-control video per condition. Each pair shares the prompt, reference frame, noisy latent, noise level, blending coefficient, decoder, and reference-latent mask. The Gaussian direction is normalized per sample to the $\ell_2$ norm of the learned direction. All 128 videos are retained and assessed by two annotators blinded to variant identity, followed by blinded adjudication.

\begin{table}[t]
\centering
\caption{Decoded adversarial counterfactuals compared with equal-norm Gaussian controls over 64 paired held-out conditions.}
\label{tab:add_gaussian}
\small
\begin{tabularx}{\linewidth}{Xccc}
\toprule
Diagnostic & Adversarial & Gaussian & Paired effect [95\% CI] \\
\midrule
Condition-preserving local physical failure & 37.5\% & 20.3\% & $+17.2$ pp $[3.1,31.3]$ \\
Global corruption / condition change & 9.4\% & 6.3\% & $+3.1$ pp $[-6.3,12.5]$ \\
Unusable & 3.1\% & 1.6\% & $+1.6$ pp $[-3.1,6.3]$ \\
\bottomrule
\end{tabularx}
\end{table}

The learned direction produces condition-preserving local failures 1.85 times as often as the matched Gaussian control (95\% CI $[1.14,3.38]$; exact McNemar $p=0.0266$). Among adversarial outputs exhibiting any failure or unusability, $75.0\%$ are targeted local physical failures. Agreement before adjudication on local physical failure versus other is $89.8\%$, with Krippendorff's $\alpha=0.752$. These results support describing the mechanism as a physical-failure-biased denoising direction at the evaluated operating point, without implying explicit spatial localization.

\subsection{Transfer to CogVideoX-5B}
\label{sec:add_cogvideox}

We repeat the complete controlled comparison on the architecturally distinct CogVideoX-5B backbone~\citep{cogvideox}. Except for each backbone's native denoising parameterization and sampler, all methods use the same preference pairs, split, conditioning inputs, training and LoRA budgets, held-out conditions, and evaluation protocol. The comparison uses 200 held-out conditions with five generations each.

\begin{table}[t]
\centering
\caption{CogVideoX-5B evaluation. Confidence intervals are condition-clustered paired bootstrap intervals for the RobotAPO Phys-S gain over each baseline.}
\label{tab:add_cogvideox}
\small
\begin{tabular}{lcccc}
\toprule
Method & Visual-S & Phys-H & Phys-S & Gain [95\% CI] \\
\midrule
SFT & $3.039\pm0.199$ & $2.980\pm0.407$ & $2.775\pm0.626$ & $+0.193$ $[0.088,0.316]$ \\
Diffusion-DPO & $3.009\pm0.104$ & $3.010\pm0.415$ & $2.824\pm0.547$ & $+0.144$ $[0.058,0.245]$ \\
RealDPO & $3.039\pm0.198$ & $2.970\pm0.399$ & $2.786\pm0.607$ & $+0.183$ $[0.081,0.299]$ \\
\RobotAPO{} & $3.008\pm0.176$ & $3.020\pm0.410$ & $2.968\pm0.294$ & -- \\
\bottomrule
\end{tabular}
\end{table}

All three paired intervals exclude zero. Together with the Wan2.2-TI2V results, this supports portability across the two evaluated video backbones while keeping the claimed scope explicit.

\subsection{Uncertainty and Execution-Relevant Metric Analysis}
\label{sec:add_statistics}

For the Wan2.2-TI2V comparison, per-video Phys-H/Phys-S standard deviations are $0.813/0.837$ for SFT, $0.866/0.893$ for Diffusion-DPO, $0.782/0.825$ for RealDPO, and $0.953/0.937$ for \RobotAPO{}. Condition-clustered paired bootstrap estimates for the RobotAPO Phys-S gain are $+0.336$ over SFT (95\% CI $[0.269,0.401]$), $+0.243$ over Diffusion-DPO ($[0.178,0.309]$), and $+0.370$ over RealDPO ($[0.306,0.435]$).

Across all 255 generated videos used for real-robot execution, the point-biserial correlation with binary full-task success is $0.775$ ($[0.721,0.822]$) for Phys-S, compared with $0.467$ ($[0.351,0.568]$) for Visual-S and $0.452$ ($[0.359,0.527]$) for Align-S. After jointly controlling for Visual-S and Align-S over the 200 held-out AgiBot conditions, \RobotAPO{} retains adjusted Phys-S gains of $+0.160$ ($[0.099,0.227]$), $+0.101$ ($[0.042,0.161]$), and $+0.223$ ($[0.161,0.291]$) over SFT, Diffusion-DPO, and RealDPO, respectively. We use Phys-S as a complementary indicator alongside complete-video inspection and robot execution rather than as a substitute for physical evaluation.

For real-robot task success, configuration-clustered paired differences are $+31.4$ percentage points over SFT ($[15.7,49.0]$), $+19.6$ over Diffusion-DPO ($[-2.0,41.2]$), $+17.6$ over RealDPO ($[-2.0,37.3]$), and $+2.0$ over LVP-14B ($[-9.8,13.7]$). Thus, the current trial budget supports the improvement over SFT; the other intervals include zero.

\section{Dataset, Implementation, and Reproducibility Details}
\label{sec:supp}

\subsection{Dataset Documentation}
\label{sec:supp_dataset_artifacts}

To make \dataset{} auditable and reproducible, each dataset version is tracked through a set of internal manifests.
The condition manifest records the AgiBot video identifier, task instruction, visual-condition path, split assignment, and prompt hash.
The generated-candidate manifest records the generator checkpoint, sampling seed, sampling hyperparameters, output path, and condition identifier for the full $40{,}000$-video candidate pool.
For the screening stage, we store the Qwen3-VL-30B-Instruct raw response, parsed quality scores, ranking position, parsing status, and filtering decision.
For the human verification stage, we store the accept/reject decision, primary failure type, optional secondary failure notes, annotator identifier, and review status.
The final preference-pair manifest contains the condition-matched tuples $(c, x^{+}, x^{-}, e^{-})$ used by training, validation, and evaluation, where $x^{+}$ is the real successful demonstration, $x^{-}$ is the failed rollout, and $e^{-}$ is the primary physical-failure label.
The evaluation manifest records held-out prompts, generated outputs from each compared method, selected LoRA scale, official VideoScore2 outputs, and robot replay records when applicable.

\subsection{Dataset Summary and Annotation Flow}
\label{sec:supp_dataset_flow}

\dataset{} is built from a $40{,}000$-video candidate pool generated by the SFT generator under $200$ task-ID conditions from the AgiBot robot-manipulation benchmark.
These conditions cover dual-arm manipulation scenarios.
The dataset is used as a training preference set rather than as a standalone benchmark with an official test split.
Evaluation is performed on disjoint held-out AgiBot task conditions, each paired with a real first-frame reference image and task instruction.
The candidate pool is not used directly as training supervision.
Instead, generated rollouts are first saved with their prompts, visual conditions, checkpoint identifiers, sampling seeds, and output paths.
Qwen3-VL-30B-Instruct is then used as a quality-control screener to rank candidates by interaction quality, 3D consistency, and instruction following.
Annotators inspect candidates in descending screening order until $10{,}000$ negatives are accepted. This requires reviewing $12{,}935$ candidates and excludes $2{,}935$ videos without a visible physical failure or sufficient evidence for reliable labeling.

Each reviewed candidate is independently annotated as a full video by two mutually blinded annotators using the unified five-category failure taxonomy in \cref{sec:supp_failure_taxonomy}, augmented with no-visible-failure and uncertain outcomes during curation.
Annotators record the most severe and visually salient visible failure as the primary label.
Their pre-adjudication seven-way judgments agree on $10{,}579/12{,}935$ candidates ($81.79\%$), with nominal Krippendorff's $\alpha=0.7832$. The remaining $2{,}356$ candidates are sent to panel adjudication before the final preference manifest is constructed.
Only human-verified condition-matched pairs are used as preference supervision, yielding the final $10{,}000$ curated samples.
Labels such as no significant physical issue and uncertain are retained for filtering and auditability, but they are not treated as physical-failure categories.
Secondary notes are also retained for auditing, while the training objective uses only the binary preference relation and the primary label for balancing and analysis.

\subsection{Qwen3-VL-30B-Instruct Screening Protocol}
\label{sec:supp_vlm_prompt}

The Qwen3-VL stage is used only for candidate quality control during dataset construction.
Official VideoScore2 is used separately for automatic evaluation in \cref{sec:exp_main}.
During screening, the implementation samples $16$ uniformly spaced RGB frames per video, conditions Qwen3-VL-30B-Instruct on the task instruction, and uses deterministic decoding with \texttt{do\_sample=False} and \texttt{max\_new\_tokens=400}.
The model is asked to evaluate only visually observable evidence in the sampled frames and to return a valid JSON object with three integer scores and concise evidence-based reasons.
The three scoring fields are \texttt{Interaction\_Quality}, \texttt{Perspectivity}, and \texttt{Instruction\_Following}.
\texttt{Interaction\_Quality} covers robot--object contact, penetration, friction, force transfer, grasping, lifting, and physically plausible manipulation.
\texttt{Perspectivity} covers 3D consistency, depth ordering, camera geometry, occlusion, scale stability, and floating artifacts.
\texttt{Instruction\_Following} compares the video against the task instruction and penalizes wrong objects, wrong actions, wrong embodiments, and human-hand hallucinations.
Candidates are ranked by the mean of the three parsed scores, and the top $10{,}000$ videos are sent to human annotation.
The recorded metadata includes the Qwen3-VL model path or version, decoding parameters, raw response, parsed scores, parsing status, aggregate rank, and filtering decision for each candidate.

\subsection{Human Verification Protocol}
\label{sec:supp_human_protocol}

Human verification is used after VLM screening to ensure that every negative training sample contains a visible manipulation-related physical failure under the matched condition.
Each candidate is independently labeled by two annotators, and disagreements are resolved through an additional review pass.
Annotators reject or mark a candidate as uncertain when the video is corrupted, too unclear to judge, mismatched with the positive sample, unrelated to manipulation physics, or ambiguous enough that it cannot be reliably labeled as success or failure.
Accepted negative samples receive one primary physical-failure label from the taxonomy in \cref{sec:supp_failure_taxonomy}.
When multiple visible failures are present, annotators choose the most severe and visually salient one and may record secondary notes for auditability.
For AgiBot preference construction, only human-verified condition-matched pairs are used as training supervision.

\subsection{Failure Type Taxonomy}
\label{sec:supp_failure_taxonomy}

\Cref{tab:supp_failure_taxonomy} restates the unified five-category taxonomy used for human annotation, training metadata, and failure-focused evaluation.
The contact-causality category is intentionally inclusive: it covers both premature object motion before effective contact and action or state changes without effective contact.

\begin{table}[!htbp]
\centering
\caption{Unified five-category failure taxonomy used for \dataset{} curation, training metadata, and failure-focused evaluation. The contact-causality category covers both premature object motion and action/state changes without effective contact.}
\label{tab:supp_failure_taxonomy}
\small
\begin{tabularx}{\linewidth}{l X X}
\toprule
\textbf{Failure Type} & \textbf{Definition} & \textbf{Typical Evidence} \\
\midrule
Object interpenetration
& Robot, object, or scene geometry passes through another entity.
& Gripper penetrates object, or object sinks into table. \\
\addlinespace[2pt]
Entity consistency error
& Robot or object identity changes in appearance, shape, or existence over time.
& Object flickers, disappears, changes shape, or changes identity across frames. \\
\addlinespace[2pt]
Contact-causality error
& Object motion or state transition is not causally supported by valid contact or manipulation, including premature motion before effective contact and action/state changes without effective contact.
& Object moves before contact, state changes without effective contact, or gripper closes while the object slides without being grasped. \\
\addlinespace[2pt]
Unsupported floating after release
& An object remains suspended after losing support.
& Released object stays in mid-air or does not settle naturally. \\
\addlinespace[2pt]
Invalid interaction / other
& No valid manipulation event under the task condition, or other visible failure not covered above.
& Robot grasps air, never interacts with the target, or executes a misaligned motion that breaks the task. \\
\bottomrule
\end{tabularx}
\end{table}

The adjudicated primary-category distribution contains 2,072 interpenetration samples (20.72\%), 1,632 entity-consistency samples (16.32\%), 2,936 contact-causality samples (29.36\%), 1,286 unsupported-floating samples (12.86\%), and 2,074 invalid-interaction/other samples (20.74\%). These mutually exclusive primary labels sum to 10,000; secondary failure tags are retained for auditing, while training uses only the condition-matched binary preference.

\subsection{Implementation Details}
\label{sec:supp_implementation}

We start from the SFT Wan2.2-TI2V flow-matching generator.
During \RobotAPO{} post-training, the SFT generator weights are frozen and two LoRA adapters are attached to the attention projections \texttt{to\_q}, \texttt{to\_k}, \texttt{to\_v}, and \texttt{to\_out.0}.
The policy adapter together with the SFT generator defines the deployed policy generator, while the adversary adapter is used only during post-training.
At inference time, the SFT generator is loaded with only the policy adapter.

Each preference pair samples synchronized frame indices from the preferred and failed videos, applies shared resize-crop transforms, and normalizes frames to the model input range.
Prompt embeddings are loaded from cache when available; otherwise they are computed with a frozen text encoder.
First-frame reference conditioning follows the Wan2.2-TI2V latent-space conditioning interface through mask-guided composition, with expanded-timestep conditioning enabled when tighter temporal conditioning is required.

Optimization uses AdamW for both policy and adversary updates, BF16 mixed precision, DeepSpeed ZeRO-2, gradient accumulation, and gradient clipping.
Training starts with a policy-only warm-up and then alternates policy and adversary phases.
The adversary is warm-started from the policy LoRA weights at the first adversary phase.
The training configuration records the pairwise temperature $\gamma$, the adversary coupling $\lambda_{\mathrm{adv}}$, and the sampling range of the hard-negative mixing weight $\lambda$ in \cref{eq:hard_pred_blend}.
The regularizer $\mathcal{L}_{\mathrm{reg}}$ is implemented as a small L2 weight-norm penalty on the trainable LoRA parameters of $f_{\pi}$.

\subsection{Compute Resources}
\label{sec:supp_compute_resources}

The main training runs used 8 NVIDIA RTX PRO 6000 GPUs for approximately 200 wall-clock hours, corresponding to about 1,600 GPU-hours.
This estimate covers the main policy and adversary training runs reported in the paper.
The Qwen3-VL screening, VideoScore2 evaluation, and robot replay stages are comparatively lighter-weight and are tracked separately in the experiment manifests.

\subsection{Official VideoScore2 Evaluation Protocol}
\label{sec:supp_vlm_eval}

For automatic video scoring, each generated video and its task prompt are evaluated with the official VideoScore2 implementation~\citep{he2025videoscore2}.
Following the released inference code, we report the three VideoScore2 dimensions: visual quality, text-to-video alignment, and physical/common-sense consistency.
For each dimension, \textbf{H} denotes the parsed hard integer score and \textbf{S} denotes the official logit-based soft score, both on a $1$--$5$ scale.
The evaluation manifest records the video path, prompt, raw VideoScore2 output, parsed hard scores, soft scores, and any parsing errors.

\subsection{Limitations and Broader Impacts}
\label{sec:supp_limitations}
\label{sec:supp_broader_impacts}

\RobotAPO{} improves the physical reliability of robot manipulation video generation by explicitly penalizing visible interaction failures under matched task conditions.
Its visually grounded preference objective sharpens the boundary between plausible rollouts and physically invalid ones, but it does not use explicit 3D geometry, force sensing, or rigid-body simulation priors.
Thus, highly complex multi-object interactions, severe occlusion, deformable-object dynamics, and closed-loop execution feedback remain outside the technical focus of this study.
The original \dataset{} pairs a real successful demonstration with a generated failure, so it is condition-matched but not source-invariant. The source-matched G+G control in \cref{sec:add_gg} reduces this concern, but it does not change the construction of the original dataset.
The demonstrated generalization scope is limited to two video backbones, held-out AgiBot conditions, and matched limited-data adaptation to the AgileX platform; it does not establish zero-shot transfer across arbitrary embodiments or task families.
In addition, \dataset{} records one primary visible-failure label for each negative example; secondary artifacts are retained as annotation notes when useful, but are not modeled as separate rewards.
Future work could incorporate richer state, depth, contact, or robot-execution signals to further improve visual-to-physical alignment.

Broadly, this work supports safer and more efficient research on embodied video generation by reducing physically implausible rollouts before robot replay or downstream execution.
More physically consistent generated plans may reduce wasted real-world trials and make model failures easier to diagnose, but they should not be treated as verified execution plans without task-specific validation, safety checks, and human oversight.
The data used in this work consists of robot manipulation videos and generated rollouts rather than personal or sensitive human data; any future release of data, annotations, or checkpoints will document intended use, known limitations, and applicable asset licenses.

\subsection{Reproducibility Checklist}
\label{sec:supp_repro}

The end-to-end pipeline can be reproduced by following the same staged construction used in our experiments.
First, one trains or loads the SFT Wan2.2-TI2V generator on AgiBot manipulation videos and generates the $40{,}000$-video candidate pool for the $200$ AgiBot benchmark task-ID conditions with recorded seeds and sampling parameters.
Second, Qwen3-VL-30B-Instruct quality-control screening is run on the generated candidates, and the shortlisted videos are independently double-annotated before disagreement review.
Third, the final train, validation, and held-out evaluation preference manifests are built with no condition leakage, yielding the $10{,}000$ curated condition-matched preference samples.
Fourth, prompt embeddings are precomputed and \RobotAPO{} training is launched with policy warm-up followed by alternating policy and adversary phases.
Finally, controlled internal baselines and ablations are trained under the same framework, and all internal checkpoints are evaluated with the official VideoScore2 protocol, a fixed shared LoRA scale of $0.4$, the self-collected real-robot replay pipeline based on frozen AnyPos action extraction, and the failure-focused qualitative comparison protocol.
LVP is evaluated only from its released checkpoint as an external embodied-video reference.

\end{document}